\documentclass[10pt,twocolumn,letterpaper]{article}

\usepackage[pagenumbers]{cvpr}              
\usepackage{multirow}
\usepackage[utf8]{inputenc} 
\usepackage[T1]{fontenc}    
\usepackage{url}            
\usepackage{booktabs}       
\usepackage{amsfonts}       
\usepackage{nicefrac}       
\usepackage{microtype}      
\usepackage{xcolor}         
\usepackage{xspace}
\usepackage{graphicx}
\usepackage{wrapfig}
\usepackage{amsmath}
\usepackage{amssymb}
\usepackage{bm}
\usepackage{tabularx}
\usepackage[linesnumbered,ruled,vlined]{algorithm2e}
\usepackage{paralist}
\usepackage{mathtools}

\makeatletter
\DeclareRobustCommand\onedot{\futurelet\@let@token\@onedot}
\def\@onedot{\ifx\@let@token.\else.\null\fi\xspace}
\def\eg{\emph{e.g}\onedot} 
\def\ie{\emph{i.e}\onedot} 
 
\def\etc{\emph{etc}\onedot} 
 
\def\etal{\emph{et al}\onedot}
\makeatother

\newcommand{\tref}[1]{Tab.~\ref{#1}}
\newcommand{\eref}[1]{Eq.~(\ref{#1})}
\newcommand{\fref}[1]{Fig.~\ref{#1}}

\newcommand{\argmax}{\operatornamewithlimits{argmax}}

\definecolor{cvprblue}{rgb}{0.21,0.49,0.74}
\usepackage[pagebackref,breaklinks,colorlinks,allcolors=cvprblue]{hyperref}
\hypersetup{
    urlcolor=magenta, 
    colorlinks=true  
}

\def\paperID{1669} 
\def\confName{CVPR}
\def\confYear{2025}

\makeatletter
\renewcommand{\@fnsymbol}[1]{}
\makeatother

\title{PIDSR: Complementary Polarized Image Demosaicing and Super-Resolution}
\author{
    Shuangfan Zhou$^1\textsuperscript{\#}$\thanks{\begin{tabular}[t]{@{}l@{}}
    \# Equal contribution. \;\;\;* Corresponding author. \\
    Code: \url{https://github.com/PRIS-CV/PIDSR}
    \end{tabular}}~
    Chu Zhou$^2\textsuperscript{\#}$~
    Youwei Lyu$^1$~
    Heng Guo$^1\textsuperscript{*}$~
    Zhanyu Ma$^1$~
    Boxin Shi$^{3,4}$~
    Imari Sato$^2$\\
    \small \textsuperscript{1} School of Artificial Intelligence, Beijing University of Posts and Telecommunications\\
    \small \textsuperscript{2} National Institute of Informatics\\
    \small \textsuperscript{3} State Key Laboratory for Multimedia Information Processing, School of Computer Science, Peking University\\ 
    \small \textsuperscript{4} National Engineering Research Center of Visual Technology, School of Computer Science, Peking University\\
    \small{\texttt{\{zhoushuangfan, youweilv, guoheng, mazhanyu\}@bupt.edu.cn~~~zhou\_chu@hotmail.com}}\\
    \small{\texttt{shiboxin@pku.edu.cn~~~imarik@nii.ac.jp}}
}

\begin{document}
\maketitle

\begin{abstract}
Polarization cameras can capture multiple polarized images with different polarizer angles in a single shot, bringing convenience to polarization-based downstream tasks. However, their direct outputs are color-polarization filter array (CPFA) raw images, requiring demosaicing to reconstruct full-resolution, full-color polarized images; unfortunately, this necessary step introduces artifacts that make polarization-related parameters such as the degree of polarization (DoP) and angle of polarization (AoP) prone to error. Besides, limited by the hardware design, the resolution of a polarization camera is often much lower than that of a conventional RGB camera. Existing polarized image demosaicing (PID) methods are limited in that they cannot enhance resolution, while polarized image super-resolution (PISR) methods, though designed to obtain high-resolution (HR) polarized images from the demosaicing results, tend to retain or even amplify errors in the DoP and AoP introduced by demosaicing artifacts. In this paper, we propose \textbf{PIDSR}, a joint framework that performs complementary \textbf{P}olarized \textbf{I}mage \textbf{D}emosaicing and \textbf{S}uper-\textbf{R}esolution, showing the ability to robustly obtain high-quality HR polarized images with more accurate DoP and AoP from a CPFA raw image in a direct manner. Experiments show our PIDSR not only achieves state-of-the-art performance on both synthetic and real data, but also facilitates downstream tasks.
\end{abstract}

\begin{figure}[t]
\centering
    \includegraphics[width=1.0\linewidth]{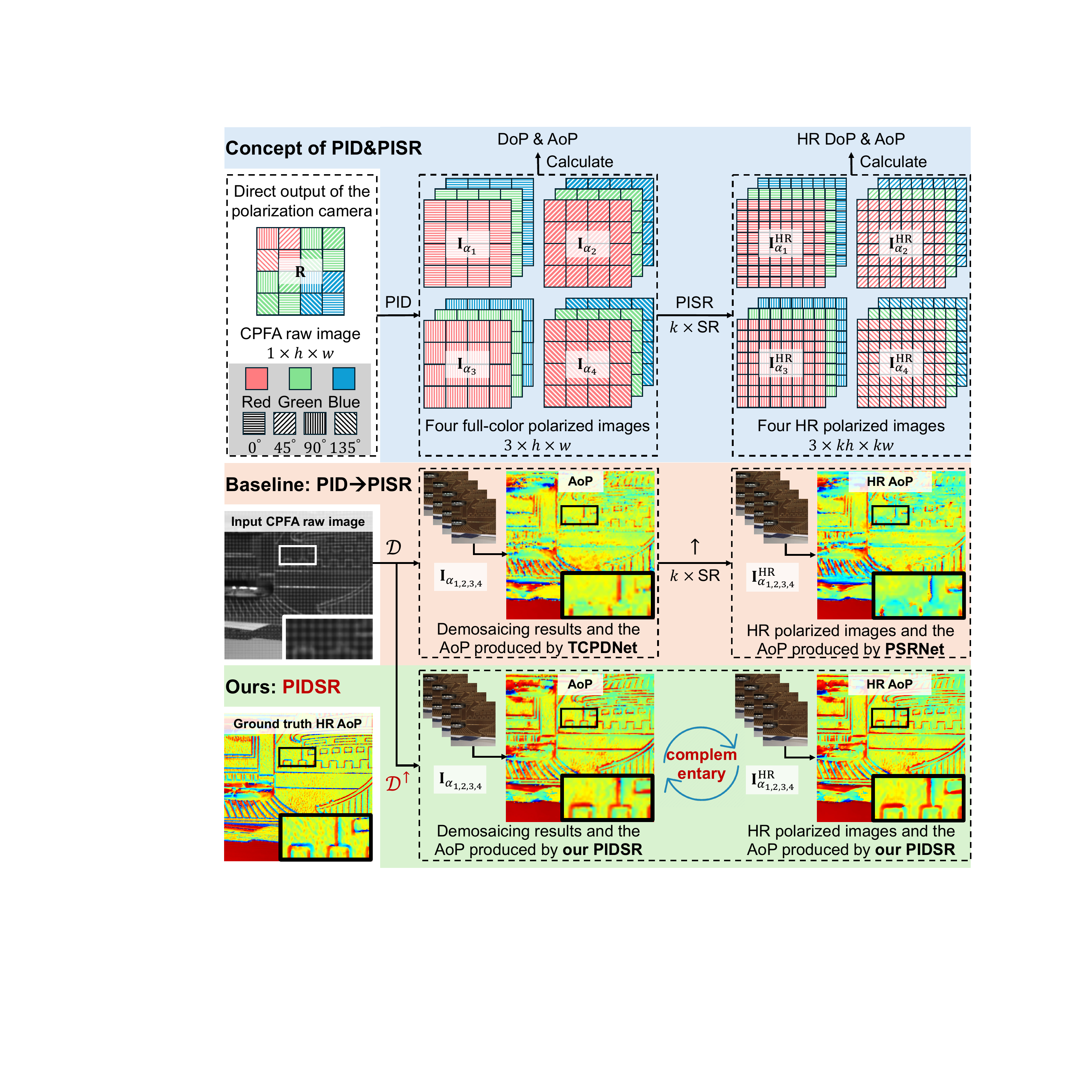}
    \caption{Top: The concept of polarized image demosaicing (PID) and polarized image super-resolution (PISR). Mid: An example shows that the baseline (PID$\rightarrow$PISR) works in a sequential manner, where the AoPs calculated from the demosaicing results (produced by TCPDNet \cite{nguyen2022two}) and the HR polarized images (produced by PSRNet \cite{hu2023polarized}) suffer from severe artifacts. Bottom: An example shows that our PIDSR works in a complementary manner, where the calculated AoPs are more accurate. We choose $k=4$ here.}
    \label{fig: Teaser}
\end{figure}

\section{Introduction}
Polarization-based vision has benefited various applications, such as shape from polarization \cite{cao2023multi, dave2022pandora}, reflection removal \cite{lyu2019reflection}, image dehazing \cite{zhou2021learning}, HDR imaging \cite{zhou2023polarizationHDR}, \etc. By fully utilizing the physical clues encoded in the polarization-relevant parameters such as the degree of polarization (DoP) and angle of polarization (AoP), polarization-based methods often achieve higher performance compared with the image-based ones, showing promising potentials. To acquire the DoP and AoP, at least three polarized images with different polarizer angles are required. While a polarizer can be used for this purpose, it demands multiple shots, making the capture process quite inconvenient. Empowered by the division of focal plane (DoFP) technology, a polarization camera can capture four color polarized images with different polarizer angles ($0^{\circ}, 45^{\circ}, 90^{\circ}, 135^{\circ}$) in a single shot, bringing convenience to the acquisition of the DoP and AoP.

Since DoFP uses color-polarization filter array (CPFA) to record the color and polarization information simultaneously, the direct output of a polarization camera is a CPFA raw image. As shown in the left part of \fref{fig: Teaser} (top), each pixel in a CPFA raw image contains information about only one color channel and one polarizer angle, which means that demosaicing is required to reconstruct the corresponding polarized images. Since unavoidable demosaicing artifacts tend to be amplified by the non-linearity of subsequent calculations, the DoP and AoP acquired from a polarization camera usually have a higher level of error than those acquired from a polarizer, making the physical clues less distinctive. Besides, limited by the hardware design, the resolution of a polarization camera is often much lower than that of a conventional RGB camera, restricting the fidelity of the recorded information. Thus, obtaining high-quality high-resolution (HR) polarized images with more accurate DoP and AoP from a polarization camera is of practical significance.

Despite its importance, a practical and reliable approach to simultaneously achieve demosaicing and super-resolution of polarization images has yet to be developed. As shown in \fref{fig: Teaser} (top), the most straightforward way is to sequentially perform polarized image demosaicing (PID) \cite{maeda2019polanalyser, morimatsu2021monochrome, nguyen2022two} and polarized image super-resolution (PISR) \cite{hu2023polarized, yu2023color} on the CPFA raw image, \ie, perform ``PID$\rightarrow$PISR''. Defining $\mathbf{R} \in \mathbb{R}^{1 \times h \times w}$ ($h$ and $w$ are the height and width respectively) as the CPFA raw image, $\mathbf{I}_{\alpha_{1,2,3,4}} \in \mathbb{R}^{3 \times h \times w}$ ($\alpha_{1,2,3,4}=0^{\circ}, 45^{\circ}, 90^{\circ}, 135^{\circ}$ are the polarizer angles) as the four full-color polarized images, and $\mathbf{I}^\text{HR}_{\alpha_{1,2,3,4}} \in \mathbb{R}^{3 \times kh \times kw}$ ($k$ is the SR scale) as the four HR polarized images respectively, the process of PID$\rightarrow$PISR can be written as
\begin{equation}
    \label{eq: PID+PISR}
     \mathbf{I}_{\alpha_{1,2,3,4}} = \mathcal{D}(\mathbf{R}) \;\text{first, then}\; \mathbf{I}^\text{HR}_{\alpha_{1,2,3,4}} = \uparrow(\mathbf{I}_{\alpha_{1,2,3,4}}),
\end{equation}
where $\mathcal{D}$ and $\uparrow$ represent demosaicing and super-resolution (SR) respectively. However, PID$\rightarrow$PISR would produce degenerated results, reducing the accuracy of the DoP and AoP, as shown in \fref{fig: Teaser} (mid). This is because existing PISR methods \cite{hu2023polarized, yu2023color} usually assume that the inputs are free of demosaicing artifacts, while existing PID methods \cite{maeda2019polanalyser, morimatsu2021monochrome, nguyen2022two} cannot guarantee perfect outputs. Therefore, the essential question needs to be addressed is: \textit{How to robustly obtain $\mathbf{I}^\text{HR}_{\alpha_{1,2,3,4}}$ from $\mathbf{R}$ in a direct manner?}

We observe that the spatial resolution often correlates negatively with the severity of demosaicing artifacts, suggesting that enhancing resolution can benefit PID, while suppressing demosaicing artifacts could, in turn, improve the performance of PISR. The observation indicates that PID and PISR may be complementary, \ie, optimizing both of them in a single framework can potentially enhance each other’s performance. This motivates us to propose \textbf{PIDSR}, a joint framework that performs complementary \textbf{P}olarized \textbf{I}mage \textbf{D}emosaicing and \textbf{S}uper-\textbf{R}esolution. As shown in \fref{fig: Teaser} (bottom), given a CPFA raw image $\mathbf{R}$, our PIDSR can not only output the demosaicing results $\mathbf{I}_{\alpha_{1,2,3,4}}$ with fewer artifacts, but also output the HR polarized images $\mathbf{I}^\text{HR}_{\alpha_{1,2,3,4}}$ with higher quality, which can be described as
\begin{equation}
    \label{eq: PIDSR}
    \mathbf{I}^\text{HR}_{\alpha_{1,2,3,4}}, \mathbf{I}_{\alpha_{1,2,3,4}} = \mathcal{D}^\uparrow(\mathbf{R}),
\end{equation}
where $\mathcal{D}^\uparrow$ denotes complementary demosaicing and SR. Here, it is non-trivial to carefully design the formulation of $\mathcal{D}^\uparrow$, since naively formulating $\mathcal{D}^\uparrow$ as a combination of $\mathcal{D}$ and $\uparrow$ would result in error accumulation. To reduce the level of error, we propose to formulate $\mathcal{D}^\uparrow$ as a series of polarized pixel reconstruction sub-problems, and introduce a two-stage pipeline to handle the intra-resolution and cross-resolution components of each sub-problem in a recurrent manner, fully exploiting the complementary aspects of $\mathcal{D}$ and $\uparrow$ to optimize each other jointly. Tailored to the pipeline, we design a neural network to explicitly inject the physical clues into both two stages to preserve the polarization properties, making full use of the Stokes-domain information of the polarized images. To summarize, this paper makes contributions by demonstrating: (1) \textbf{PIDSR, a complementary polarized image demosaicing and super-resolution framework}, including: (2) \textbf{a two-stage recurrent pipeline} to fundamentally reduce the level of error; and (3) \textbf{a Stokes-aided neural network} to preserve the polarization properties.

\section{Related work}
\noindent\textbf{Polarized image demosaicing (PID).} Unlike the demosaicing methods designed for conventional RGB images \cite{hou2024dtdemo, qian2019rethink, kokkinos2018iterative} that handle the mosaic generated from color filter array (CFA), the methods designed for PID aim to deal with the mosaic from color-polarization filter array (CPFA). Maeda developed Polanalyser \cite{maeda2019polanalyser}, an open-source software that provides an interpolation-based basic PID tool, which is widely adopted in polarization-based vision tasks \cite{cao2023multi, dave2022pandora}. For higher performance, some methods attempted to adopt numerical optimization based on handcrafted priors to suppress the demosaicing artifacts \cite{liu2020new, morimatsu2020monochrome, morimatsu2021monochrome, qiu2021linear, wu2021polarization, xin2023demosaicking, lu2024hybrid, hagen2024fourier, yi2024demosaicking}. Some works adopted learning-based approaches to solve this challenging problem, including convolutional neural network (CNN) \cite{wen2019convolutional, zeng2019end, sun2021color, liu2022enhanced, nguyen2022two, lu2024polarization}, generative adversarial network (GAN) \cite{guo2024attention}, dictionary learning \cite{wen2021sparse, zhang2021polarization, luo2023sparse, luo2024learning}, \etc. Li \etal \cite{li2021no} proposed a no-reference physics-based quality assessment metric and show that it can be used to address the PID problem. \textit{However, these methods can only restore full-color polarized images from a CPFA raw image, and cannot further enhance the resolution.}

\noindent\textbf{Polarized image super-resolution (PISR).} Unlike the SR methods designed for conventional RGB images \cite{saharia2022image, wang2023omni, gao2023implicit, wang2024sinsr} that focus solely on resolution enhancement, the methods designed for PISR aim to not only improve the resolution of multiple polarized images but also preserve polarization properties in them, making the task much more challenging. Hu \etal \cite{hu2023polarized} proposed two polarized image degradation models to simulate real image degradation, and designed a network named PSRNet to perform polarization-aware SR on monochrome polarized images along with a loss function to refine the DoP and AoP in a direct manner. Yu \etal \cite{yu2023color} proposed a network named CPSRNet to perform polarization-aware SR on color polarized images, which incorporated a cross-branch activation module (CBAM) \cite{woo2018cbam} to leverage high-frequency information contained in the DoP and AoP for preserving the polarization properties explicitly. \textit{However, they are largely based on an assumption that the demosaicing artifacts are not that significant, and ignore the errors in the DoP and AoP of the input polarized images.}

\section{Method}
\subsection{Background}
\noindent\textbf{CPFA raw image formation model.} 
Since polarization cameras have a linear camera response function (\ie, the pixel values linearly relate to the input irradiance), here we follow other works \cite{lyu2019reflection, zhou2021learning} by not applying any special adjustments for non-linearity. As shown in \fref{fig: Teaser} (top), a full-color polarized image $\mathbf{I}_{\alpha_i} \in \mathbb{R}^{3 \times h \times w}$ can be regarded as a collection of single-channel polarized images, \ie, $\mathbf{I}_{\alpha_i} = \{\mathbf{I}_{\alpha_i}^{c_j}\}$, where $i=1,2,3,4$ and $\alpha_{1,2,3,4}=0^{\circ}, 45^{\circ}, 90^{\circ}, 135^{\circ}$ denote the polarizer angles, $j=r,g,b$ and $c_{r,g,b}$ denote RGB color channels. Here, each single-channel polarized image $\mathbf{I}_{\alpha_i}^{c_j}  \in \mathbb{R}^{1 \times h \times w}$ can be written as
\begin{equation}
    \label{eq: PolarizedImage}
    \mathbf{I}_{\alpha_i}^{c_j} = \mathcal{C}_j(\mathcal{P}_i(\mathbf{E})), 
\end{equation}
where $\mathbf{E} \in \mathbb{R}^{1 \times h \times w}$ denotes the input irradiance sampled by an $h \times w$ pixel array, $\mathcal{C}_j$ and $\mathcal{P}_i$ denote the color and polarization filtering operations at $c_j$ and $\alpha_i$ performed by the CPFA respectively. A CPFA raw image $\mathbf{R} \in \mathbb{R}^{3 \times h \times w}$ captured by a polarization camera can be regarded as the weighted sum of each single-channel polarized image $\mathbf{I}_{\alpha_i}^{c_j}$:
\begin{equation}
    \label{eq: CPFA}
        \mathbf{R} = \sum_{\substack{i \in \{1,2,3,4\} \\ j \in \{r,g,b\}}} \mathbf{M}_{ij} \cdot \mathbf{I}_{\alpha_i}^{c_j} = \sum_{\substack{i \in \{1,2,3,4\} \\ j \in \{r,g,b\}}} \mathbf{M}_{ij} \cdot (\mathcal{C}_j(\mathcal{P}_i(\mathbf{E}))),
\end{equation}
where $\mathbf{M}_{ij} \in \mathbb{R}^{1 \times h \times w}$ denotes the weight of each summation term whose pixel value at coordinates $(x, y)$ satisfies
\begin{equation}
    \label{eq: Mask}
    \mathbf{M}_{ij}(x, y) = 
    \begin{cases}
        1 & \text{if $\mathbf{I}_{\alpha_i}^{c_j}(x, y)$ is in the CPFA pattern}\\
        0 & \text{otherwise}\\
    \end{cases}
    .
\end{equation}
Combing \eref{eq: PolarizedImage} and \eref{eq: CPFA}, we can see that PID is similar to performing interpolation on the missing pixels (\ie, the pixels at coordinates $(x, y)$ satisfying $\mathbf{M}_{ij}(x, y)=0$) from one out of twelve necessary intensity measurements, and it is an ill-posed problem without closed form solution.

\noindent\textbf{Acquisition of the DoP and AoP.} Given a CPFA raw image $\mathbf{R}$, one can perform PID on it to obtain four full-color polarized images $\mathbf{I}_{\alpha_{1,2,3,4}}$ and use them to acquire the DoP $\mathbf{p} \in [0,1]$ and AoP $\bm{\theta} \in [0, \pi]$ for downstream tasks by 
\begin{equation}
    \label{eq: DoPAoP}
    \mathbf{p}=\frac{\sqrt{\mathbf{S}_1^2 + \mathbf{S}_2^2}}{\mathbf{S}_0} \;\;\text{and}\;\; \bm{\theta}=\frac{1}{2}\arctan(\frac{\mathbf{S}_2}{\mathbf{S}_1}),
\end{equation} 
where $\mathbf{S}_{0,1,2}$\footnote{$\mathbf{S}_0$ describes the total intensity (which can be regarded as the unpolarized image), and $\mathbf{S}_1$ ($\mathbf{S}_2$) describes the difference between the intensity of the vertical and horizontal ($135^{\circ}$ and $45^{\circ}$) polarized light.} are called the Stokes parameters \cite{hecht2012optics, konnen1985polarized} that can be computed as
\begin{equation}
    \label{eq: StokesParameters}
    \begin{dcases}
    \mathbf{S}_0 = 2\bar{\mathbf{I}}_{\alpha_i} = \mathbf{I}_{\alpha_1} + \mathbf{I}_{\alpha_3} = \mathbf{I}_{\alpha_2} + \mathbf{I}_{\alpha_4}\\
    \mathbf{S}_1 = \mathbf{I}_{\alpha_3} - \mathbf{I}_{\alpha_1}, \;\;\text{and}\;\; \mathbf{S}_2 = \mathbf{I}_{\alpha_4} - \mathbf{I}_{\alpha_2}
    \end{dcases}
    ,
\end{equation}
where $\bar{\mathbf{I}}_{\alpha_i} = \sum_{i=1}^4 \mathbf{I}_{\alpha_i} / 4$ is the average polarized image. 

\noindent\textbf{Acquisition of the HR counterparts.} As the polarized images $\mathbf{I}_{\alpha_{1,2,3,4}}$ become available, one can perform PISR on them to acquire their HR counterparts $\mathbf{I}^\text{HR}_{\alpha_{1,2,3,4}}$. Similarly, the HR counterparts of the Stokes parameters $\mathbf{S}^\text{HR}_{0,1,2}$, DoP $\mathbf{p}^\text{HR}$ and AoP $\bm{\theta}^\text{HR}$ can also be acquired by substituting $\mathbf{I}_{\alpha_{1,2,3,4}}$ with $\mathbf{I}^\text{HR}_{\alpha_{1,2,3,4}}$ in \eref{eq: StokesParameters} and \eref{eq: DoPAoP}. It is important to note existing PISR methods \cite{hu2023polarized, yu2023color} cannot directly perform super-resolution on CPFA raw images, and they require PID as a pre-processing step to generate the polarized images first.

\subsection{Motivation and overall framework}
\begin{figure}[t]
\centering
    \includegraphics[width=1.0\linewidth]{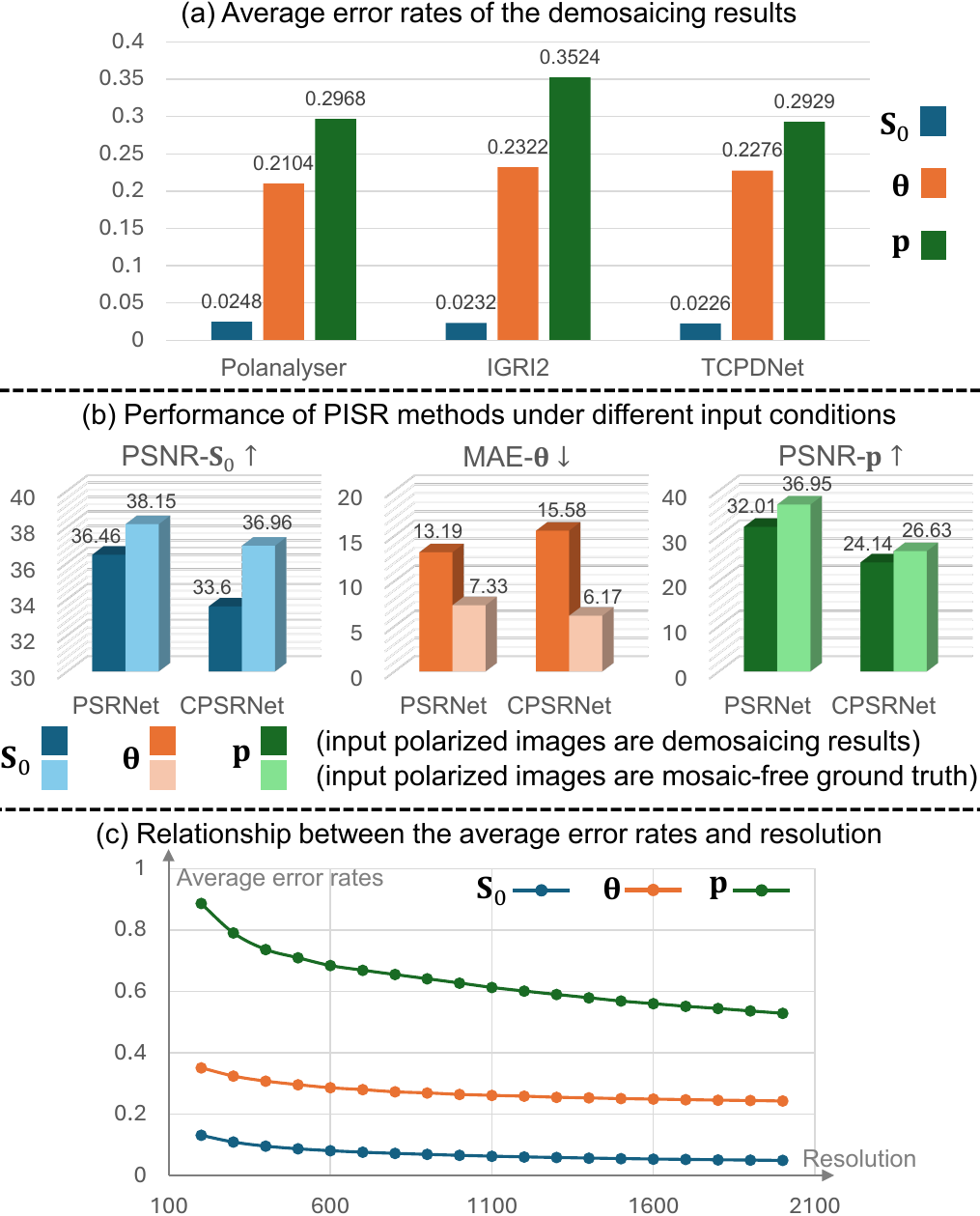}
    \caption{(a) The average error rates of $\mathbf{p}$ and $\bm{\theta}$ are much larger than the one of $\mathbf{S}_0$ for PID methods (Polanalyser \cite{maeda2019polanalyser}, IGRI2 \cite{morimatsu2021monochrome}, and TCPDNet \cite{nguyen2022two}). (b) The performance of PISR methods (PSRNet \cite{hu2023polarized} and CPSRNet \cite{yu2023color}) on both $\mathbf{p}$, $\bm{\theta}$, and $\mathbf{S}_0$ is much better when using mosaic-free ground truth polarized images compared with using demosaicing results. (c) The average error rates of both $\mathbf{p}$, $\bm{\theta}$, and $\mathbf{S}_0$ decrease as the resolution increases.}
    \label{fig: Verification}
\end{figure}

As indicated in \eref{eq: StokesParameters} and \eref{eq: DoPAoP}, $\mathbf{p}$ and $\bm{\theta}$ exhibit non-linear relationships with $\mathbf{I}_{\alpha_{1,2,3,4}}$. This non-linearity would exacerbate demosaicing artifacts, meaning errors arising from imperfections in PID (\eg, inaccurate interpolation, failure to handle sensor noise, \etc) are more noticeable in $\mathbf{p}$ and $\bm{\theta}$ than in $\mathbf{I}_{\alpha_{1,2,3,4}}$. To verify it, we design an experiment on our test dataset to evaluate the average error rates of $\mathbf{p}$, $\bm{\theta}$, and $\mathbf{S}_0$\footnote{Since $\mathbf{S}_0$ has a linear relationship with $\mathbf{I}_{\alpha_{1,2,3,4}}$ (see \eref{eq: StokesParameters}), we can use $\mathbf{S}_0$ to represent $\mathbf{I}_{\alpha_{1,2,3,4}}$ in such a proof-of-concept experiment.} acquired from the demosaicing results. Here, we choose Polanalyser \cite{maeda2019polanalyser}, IGRI2 \cite{morimatsu2021monochrome}, and TCPDNet \cite{nguyen2022two} as the PID methods, and define the error rate of a variable $\mathbf{v}$ (normalized to $[0, 1]$) similar to the one in \cite{zhou2023polarization}: $\text{ER}_{\mathbf{v}} = \frac{\sum_{\text{p/w}}|\mathbf{v}-\mathbf{v}_\textbf{gt}|}{\sum_{\text{p/w}}\mathbf{v}}$, where $\sum_{\text{p/w}}$ denotes the pixel-wise sum, the subscript gt stands for the ground truth throughout this paper. As shown in \fref{fig: Verification} (a), the average error rates of $\mathbf{p}$ and $\bm{\theta}$ are much larger than $\mathbf{S}_0$ for all PID methods. Besides, obtaining high-quality HR polarized images is challenging because the performance of PISR is constrained by the effectiveness of the pre-processing step, PID. To verify it, we design another experiment on our test dataset to evaluate the performance of two PISR methods (PSRNet \cite{hu2023polarized} and CPSRNet \cite{yu2023color}) under different input conditions. Specifically, we use polarized images generated by an existing PID method (Polanalyser \cite{maeda2019polanalyser}) and their corresponding mosaic-free ground truth as inputs respectively for comparison. As shown in \fref{fig: Verification} (b), the performance of PISR methods using polarized images generated by Polanalyser \cite{maeda2019polanalyser} as input is inferior to that using the ground truth as input for both $\mathbf{p}$, $\bm{\theta}$, and $\mathbf{S}_0$.

Notably, we observe that the spatial resolution at which the input irradiance $\mathbf{E}$ is sampled (\ie, the same scene sampled at different $h \times w$ resolutions) often correlates negatively with the severity of demosaicing artifacts. As a proof of concept, we use a virtual camera with varying resolutions to sample the input irradiance from rendered scenes (using Mitsuba 3\footnote{https://www.mitsuba-renderer.org/}), and adopt \eref{eq: CPFA} to obtain the CPFA raw images at different resolutions; then, a PID method (Polanalyser \cite{maeda2019polanalyser}) is adopted to produce the corresponding demosaicing results. Results are shown in \fref{fig: Verification} (c), which demonstrate the severity of demosaicing artifacts (measured using average error rates of both $\mathbf{p}$, $\bm{\theta}$, and $\mathbf{S}_0$) decreases as the resolution increases. We can see these results align with the fact that existing PID methods \cite{maeda2019polanalyser, morimatsu2021monochrome, nguyen2022two} typically leverage interactions among neighboring pixels for interpolation, where preserving finer details could lead to more accurate interpolated pixel values (\ie, at higher resolutions, artifacts such as blurring or jagged edges (aliasing) are less likely to occur). This observation suggests enhancing resolution can benefit PID. Combining the fact that suppressing demosaicing artifacts could improve the performance of PISR, we can deduce that PID and PISR may be complementary.

Based on the above analysis, we propose to design a joint framework that performs complementary polarized image demosaicing and super-resolution, named PIDSR. As shown in \eref{eq: PIDSR}, given a CPFA raw image $\mathbf{R}$ as input, our PIDSR aims to perform complementary demosaicing and SR ($\mathcal{D}^\uparrow$) on it to output not only demosaicing results $\mathbf{I}_{\alpha_{1,2,3,4}}$ but also HR polarized images $\mathbf{I}^\text{HR}_{\alpha_{1,2,3,4}}$ with more accurate DoP and AoP. Thus, the overall process of PIDSR can be regarded as maximizing a posteriori of the outputs $\mathbf{I}_{\alpha_{1,2,3,4}}$ and $\mathbf{I}^\text{HR}_{\alpha_{1,2,3,4}}$ conditioned on the inputs $\mathbf{R}$ along with the complementary demosaicing and SR function $\mathcal{D}^\uparrow$ parameterized by $\Psi$:
\begin{equation}
    \label{eq: Posteriori}
    \argmax_\Psi \mathcal{D}^\uparrow(\mathbf{I}_{\alpha_{1,2,3,4}}, \mathbf{I}^\text{HR}_{\alpha_{1,2,3,4}} | \mathbf{R}, \Psi).
\end{equation}

\begin{figure}[t]
\centering
    \includegraphics[width=1.0\linewidth]{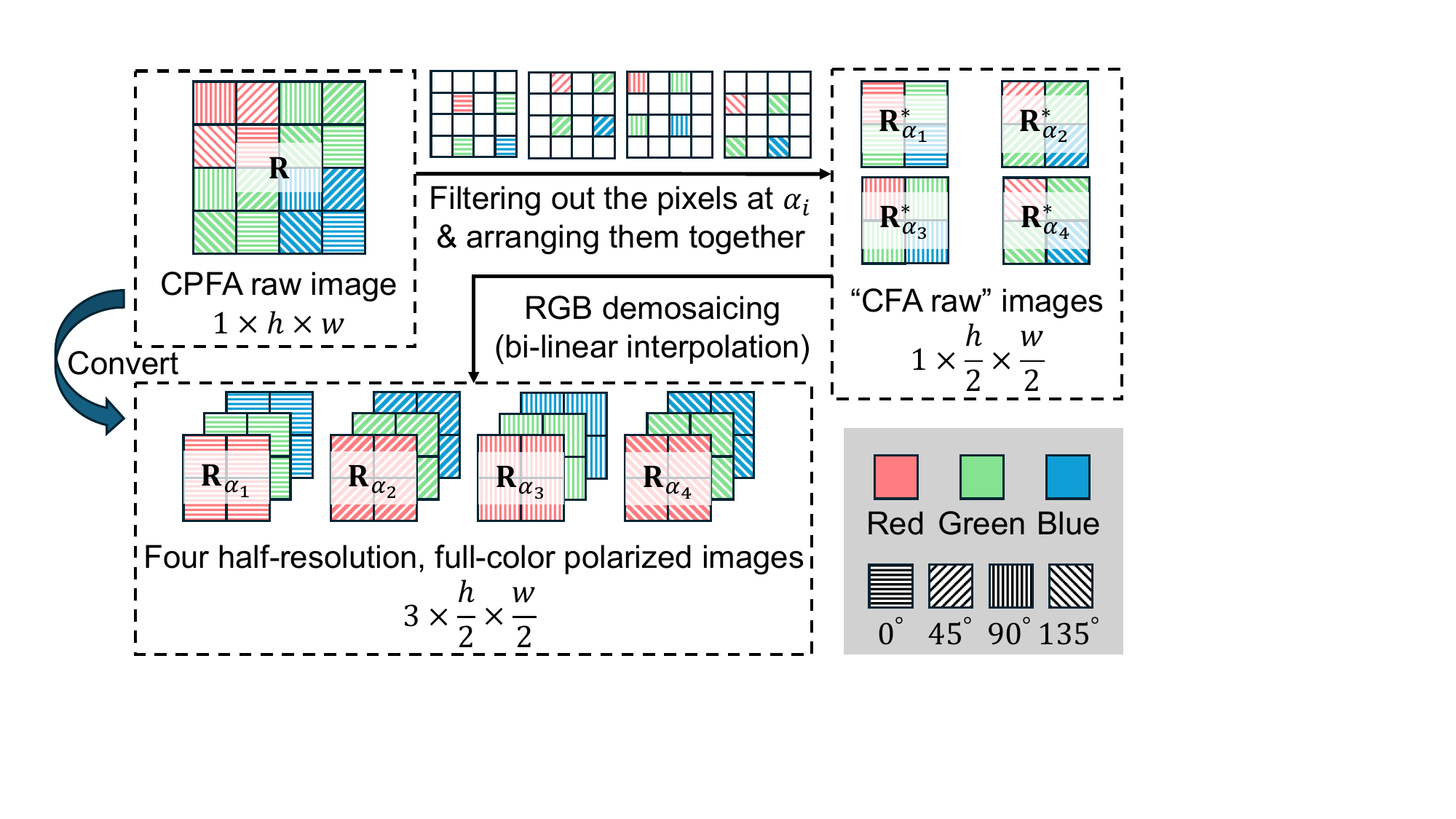}
    \caption{A CPFA raw image $\mathbf{R}$ can be approximately converted to four half-resolution, full-color polarized image $\mathbf{R}_{\alpha_{1,2,3,4}}$.}
    \label{fig: Conversion}
\end{figure}

\begin{figure*}[t]
\centering
    \includegraphics[width=1.0\linewidth]{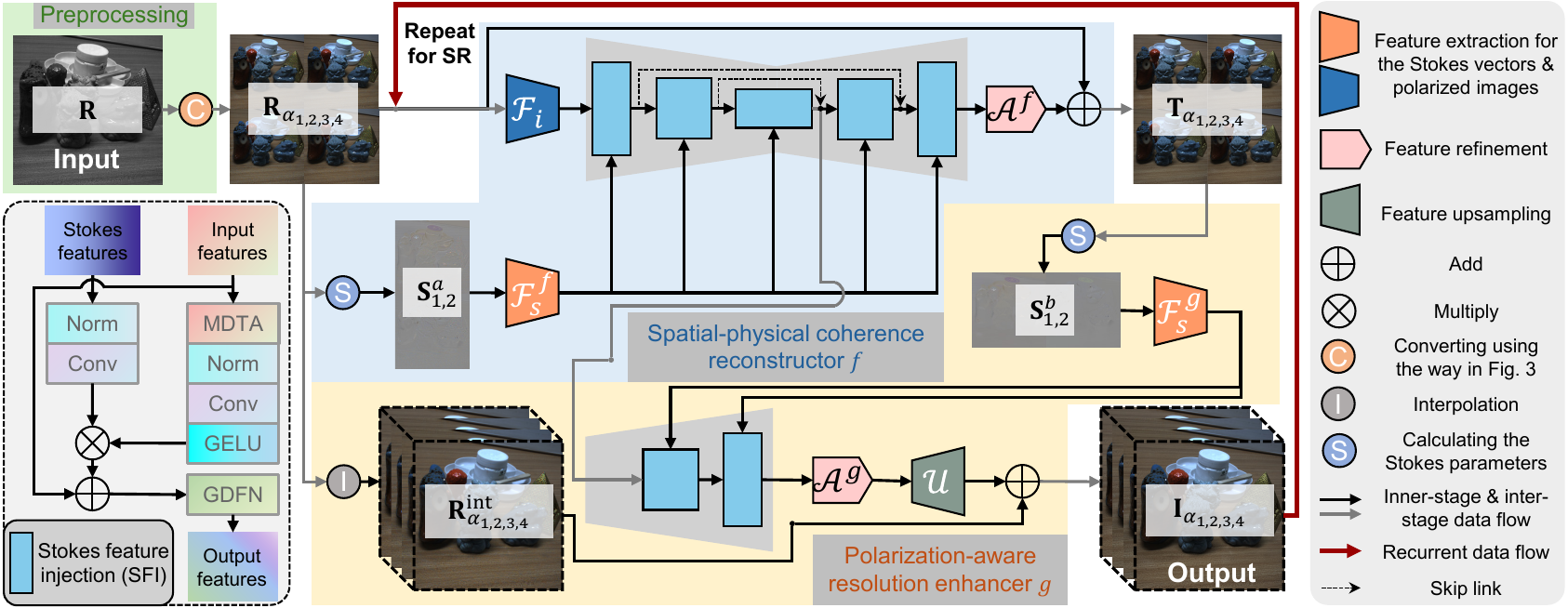}
    \caption{The workflow and network design of our PIDSR framework, consisting of two stages: spatial-physical coherence reconstructor $f(\cdot)$ and polarization-aware resolution enhancer $g(\cdot)$. Here we only illustrate the demosaicing workflow, and the SR one is in a repeated manner.}
    \label{fig: Network}
\end{figure*}

\subsection{Two-stage recurrent PIDSR pipeline}
The most straightforward way to solve the maximum a posteriori estimation problem in \eref{eq: Posteriori} is directly formulating the complementary demosaicing and SR function $\mathcal{D}^\uparrow$ as a cascade of two stages, \ie, demosaicing function $\mathcal{D}$ and SR function $\uparrow$, as shown in \eref{eq: PID+PISR}. However, such a pipeline has two main drawbacks that limit its overall performance. First, errors tend to accumulate across stages because $\mathcal{D}$ and $\uparrow$ are independent, preventing the formation of negative feedback loops to stabilize the level of error. Second, the sequential nature of these stages fails to leverage the complementary aspects of $\mathcal{D}$ and $\uparrow$, missing the opportunity for joint optimization. Therefore, a more robust pipeline is required.

Prominently, we found that a CPFA raw image $\mathbf{R}$ can be approximately converted to four half-resolution, full-color polarized images $\mathbf{R}_{\alpha_{1,2,3,4}}$. As shown in \fref{fig: Conversion}, first, by filtering out the pixels corresponding to a specific polarizer angle $\alpha_i$ from a given CPFA raw image $\mathbf{R}$ and arranging them together, we could form an image $\mathbf{R}_{\alpha_i}^* \in \mathbb{R}^{1 \times h/2 \times w/2}$ with a format similar to a CFA raw image (\ie, the mosaic pattern produced by a color filter array); then, by applying a simple RGB demosaicing method (\eg, bi-linear interpolation) to $\mathbf{R}_{\alpha_i}^*$, we could generate a half-resolution, full-color polarized image $\mathbf{R}_{\alpha_i}$, though it would exhibit spatial discontinuities between neighboring pixels. This suggests that $\mathcal{D}$ can be formulated into two sub-problems: spatial discontinuity alleviation and resolution enhancement, equivalent to performing intra-resolution and cross-resolution polarized pixel reconstruction. Similarly, $\uparrow$ can also be formulated into two sub-problems: physical correlation restoration (the intra-resolution one) and resolution enhancement (the cross-resolution one). This decoupled formulation ensures that the disrupted physical correlation among multiple polarized images, caused by demosaicing artifacts from the pre-processing step (PID), has minimal negative impact on resolution enhancement, facilitating the accurate acquisition of DoP and AoP. To this end, we could unify $\mathcal{D}$ and $\uparrow$ into a recurrent structure, which can not only avoid error accumulation but also make full use of the complementary aspects of $\mathcal{D}$ and $\uparrow$ to optimize each other jointly.

We design a two-stage recurrent PIDSR pipeline to implement $\mathcal{D}^\uparrow$, as illustrated in \fref{fig: Network}. The first stage $f$ is a \textit{spatial-physical coherence reconstructor} that performs intra-resolution pixel reconstruction, aiming to alleviate the spatial discontinuities between neighboring pixels, restore the physical correlation among multiple polarized images, and deal with the potential sensor noise; the second stage $g$ is a \textit{polarization-aware resolution enhancer} that performs cross-resolution pixel reconstruction, with a focus on both SR and preserving polarization properties. Starting with a CPFA raw image $\mathbf{R}$ as the initial input, we first approximately convert it into four half-resolution, full-color polarized images $\mathbf{R}_{\alpha_{1,2,3,4}}$ using the way shown in \fref{fig: Conversion} as a pre-processing step, then send $\mathbf{R}_{\alpha_{1,2,3,4}}$ to $f$ and $g$ in a sequential manner to finish the first round of iteration to obtain four full-color polarized images $\mathbf{I}_{\alpha_{1,2,3,4}}$; after that, we can repeat the iteration for $n$ additional rounds to produce four HR polarized images $\mathbf{I}^\text{HR}_{\alpha_{1,2,3,4}}$ with an SR factor of $k=2^n \times$.

\begin{table*}[t]
    \centering
    \caption{Quantitative comparisons on synthetic data. The comparisons involve our PIDSR, three state-of-the-art PID methods (Polanalyser \cite{maeda2019polanalyser}, IGRI2 \cite{morimatsu2021monochrome}, and TCPDNet \cite{nguyen2022two}), and the only existing two PISR methods (PSRNet \cite{hu2023polarized} and CPSRNet \cite{yu2023color}).}
    \label{tab: SyntheticData}
    \resizebox{1.\textwidth}{!}{
    \begin{tabular}{lcccccccc}
        \toprule
         {Metric} & \multicolumn{6}{c}{PSNR$\uparrow$/SSIM$\uparrow$}  & {MAE$\downarrow$}\\
        \cmidrule[0.5pt](rl){1-1}
        \cmidrule[0.5pt](rl){2-7}
        \cmidrule[0.5pt](rl){8-8}
         {\large Demosaicing} & {$\mathbf{I}_{\alpha_1} (0^{\circ})$} & {$\mathbf{I}_{\alpha_2} (45^{\circ})$} & {$\mathbf{I}_{\alpha_3} (90^{\circ})$} & {$\mathbf{I}_{\alpha_4} (135^{\circ})$} & {$\mathbf{S}_0$} & {$\mathbf{p}$}  & {$\bm{\theta}$}\\
        \midrule
        Polanalyser \cite{maeda2019polanalyser} & 31.95/0.8955 & 32.08/0.8968 & 32.44/0.8989 & 32.14/0.8973 & 33.28/0.9138 & 26.68/0.7164 & 17.8666 \\
        IGRI2 \cite{morimatsu2021monochrome} &  34.25/0.9349 & 34.33/0.9359 & 34.64/0.9369 & 34.46/0.9365 & 35.50/0.9450 & 27.78/0.7486 & 16.5830 \\
        TCPDNet \cite{nguyen2022two} & 37.26/0.9585 & 37.60/0.9602 & 37.90/0.9609 & 37.81/0.9612 & 38.65/0.9669 & 32.26/0.8262 & 13.1881\\
        \textbf{PIDSR} & \textbf{38.90/0.9717} & \textbf{38.98/0.9719} & \textbf{39.11/0.9721} & \textbf{39.21/0.9732} & \textbf{40.24/0.9780} & \textbf{33.33/0.8447} & \textbf{12.2383} \\
        \bottomrule
        \toprule
         {\large Super-resolution} & {$\mathbf{I}^\text{HR}_{\alpha_1} (0^{\circ})$} & {$\mathbf{I}^\text{HR}_{\alpha_2} (45^{\circ})$} & {$\mathbf{I}^\text{HR}_{\alpha_3} (90^{\circ})$} & {$\mathbf{I}^\text{HR}_{\alpha_4} (135^{\circ})$} & {$\mathbf{S}_0^\text{HR}$} & {$\mathbf{p}^\text{HR}$}  & {$\bm{\theta}^\text{HR}$}\\
        \midrule
        PSRNet (2$\times$) \cite{hu2023polarized} & 35.66/0.9309 & 35.49/0.9301 & 35.65/0.9306 & 35.65/0.9319 & 36.46/0.9439 & 32.01/0.8298 & 13.1884 \\
        CPSRNet (2$\times$) \cite{yu2023color} & 32.97/0.8936 & 33.11/0.8944 & 33.35/0.8947 & 33.17/0.8958 & 33.60/0.9021 & 24.14/0.7649 & 15.5811 \\
        \textbf{PIDSR} (2$\times$) & \textbf{36.55/0.9488} & \textbf{36.64/0.9493} & \textbf{36.85/0.9502} & \textbf{36.77/0.9505} & \textbf{37.44/0.9553} & \textbf{32.97/0.8438} & \textbf{12.3520} \\
        \midrule
        PSRNet (4$\times$) \cite{hu2023polarized} & 35.15/0.9227 & 35.41/0.9247 & 35.74/0.9264 & 35.57/0.9257 & 36.13/0.9311 & 31.95/0.8305 & 13.7751 \\
        CPSRNet (4$\times$) \cite{yu2023color} & 30.82/0.8599 & 30.75/0.8596 & 30.98/0.8600 & 30.76/0.8608 & 31.16/0.8677 & 22.52/0.7325 & 16.5469 \\
        \textbf{PIDSR} (4$\times$) & \textbf{35.48/0.9297} & \textbf{35.58/0.9307} & \textbf{35.83/0.9321} & \textbf{35.70/0.9319} & \textbf{36.31/0.9371} & \textbf{32.43/0.8379} & \textbf{13.0520}  \\
        \bottomrule
        \end{tabular}
    }
\end{table*}

\subsection{Stokes-aided PIDSR network}
\noindent\textbf{Spatial-physical coherence reconstructor ($f$).} As shown in the first stage of \fref{fig: Network}, it aims to solve the intra-resolution polarized pixel reconstruction sub-problem, which alleviates the inherent spatial discontinuity in $\mathbf{R}_{\alpha_{1,2,3,4}}$ and restores the imperfect physical correlation in $\mathbf{I}_{\alpha_{1,2,3,4}}$ during the demosaicing and SR workflows, respectively. Taking the demosaicing workflow as an example, this stage learns the residual between $\mathbf{R}_{\alpha_{1,2,3,4}}$ and $\mathbf{T}_{\alpha_{1,2,3,4}}$ (which are the spatially continuous intermediate results). First, two feature extraction heads $\mathcal{F}_i$ and $\mathcal{F}_s^f$ are used to extract the image and polarization features from $\mathbf{R}_{\alpha_{1,2,3,4}}$ and their corresponding Stokes parameters $\mathbf{S}_{1,2}^a$ respectively. Then, a backbone network is adopted to process the extracted features to compensate the missing spatial information in the high-dimensional feature space. Here, we should not directly concatenate the extracted features and send them into the backbone network, since the domain gap between the features of $\mathbf{R}_{\alpha_{1,2,3,4}}$ and $\mathbf{S}_{1,2}^a$ could be very large, \ie, the features of $\mathbf{R}_{\alpha_{1,2,3,4}}$ contain mainly low-frequency structures, while the features of $\mathbf{S}_{1,2}^a$ contain mainly high-frequency structures. To handle this issue, we design the backbone network as a modified U-Net \cite{ronneberger2015u} architecture, where in each scale the original convolution block is substituted with a Stokes feature injection (SFI) block to explicitly utilize the physical clues encoded in the Stokes parameters to provide guidance for bridging the domain gap. The SFI block contains two different branches for processing the input and Stokes features respectively, which learn a bias by multiplying the processed features to adjust the input features. To effectively capture long-range feature interactions, we design the SFI block to incorporate a multi-Dconv head transposed attention (MDTA) module \cite{zamir2022restormer} at the beginning of the branch for input features along with a gated-Dconv feed-forward
network (GDFN) module \cite{zamir2022restormer} before output. After the backbone network, a feature refinement block $\mathcal{A}^f$ (containing an MDTA and a GDFN module \cite{zamir2022restormer}) is used to reconstruct the residual between $\mathbf{R}_{\alpha_{1,2,3,4}}$ and $\mathbf{T}_{\alpha_{1,2,3,4}}$.

\noindent\textbf{Polarization-aware resolution enhancer ($g$).} As shown in the second stage of \fref{fig: Network}, it aims to solve the cross-resolution polarized pixel reconstruction sub-problem, which focuses on resolution enhancement during both the demosaicing and SR workflows. Also taking the demosaicing workflow as an example, this stage learns the residual between $\mathbf{R}_{\alpha_{1,2,3,4}}^{\text{int}}$ (the interpolated version of $\mathbf{R}_{\alpha_{1,2,3,4}}$) and $\mathbf{I}_{\alpha_{1,2,3,4}}$. Since $\mathbf{T}_{\alpha_{1,2,3,4}}$ are spatially continuous, their corresponding Stokes parameters $\mathbf{S}_{1,2}^b$ could offer robust physical clues to facilitate the SR process with polarization-awareness. Besides, since the backbone network in $f$ already encodes fine-grained multiscale features in the image domain, we do not need to extract features from $\mathbf{T}_{\alpha_{1,2,3,4}}$ additionally. Therefore, in this stage, we choose to directly grab the features from the coarsest level of the backbone network in $f$ and send them into a decoder (sharing the same architecture with the decoder part of the backbone network in $f$), under the guidance of the features of $\mathbf{S}_{1,2}^b$ extracted by another feature extraction head $\mathcal{F}_s^g$. Then, the output features of the decoder are fed into another feature refinement block $\mathcal{A}^g$ and a feature upsampling block $\mathcal{U}$ in a sequential manner to form the residual between $\mathbf{R}_{\alpha_{1,2,3,4}}^{\text{int}}$ and $\mathbf{I}_{\alpha_{1,2,3,4}}$.

\begin{figure*}[t]
\centering
    \includegraphics[width=\textwidth]{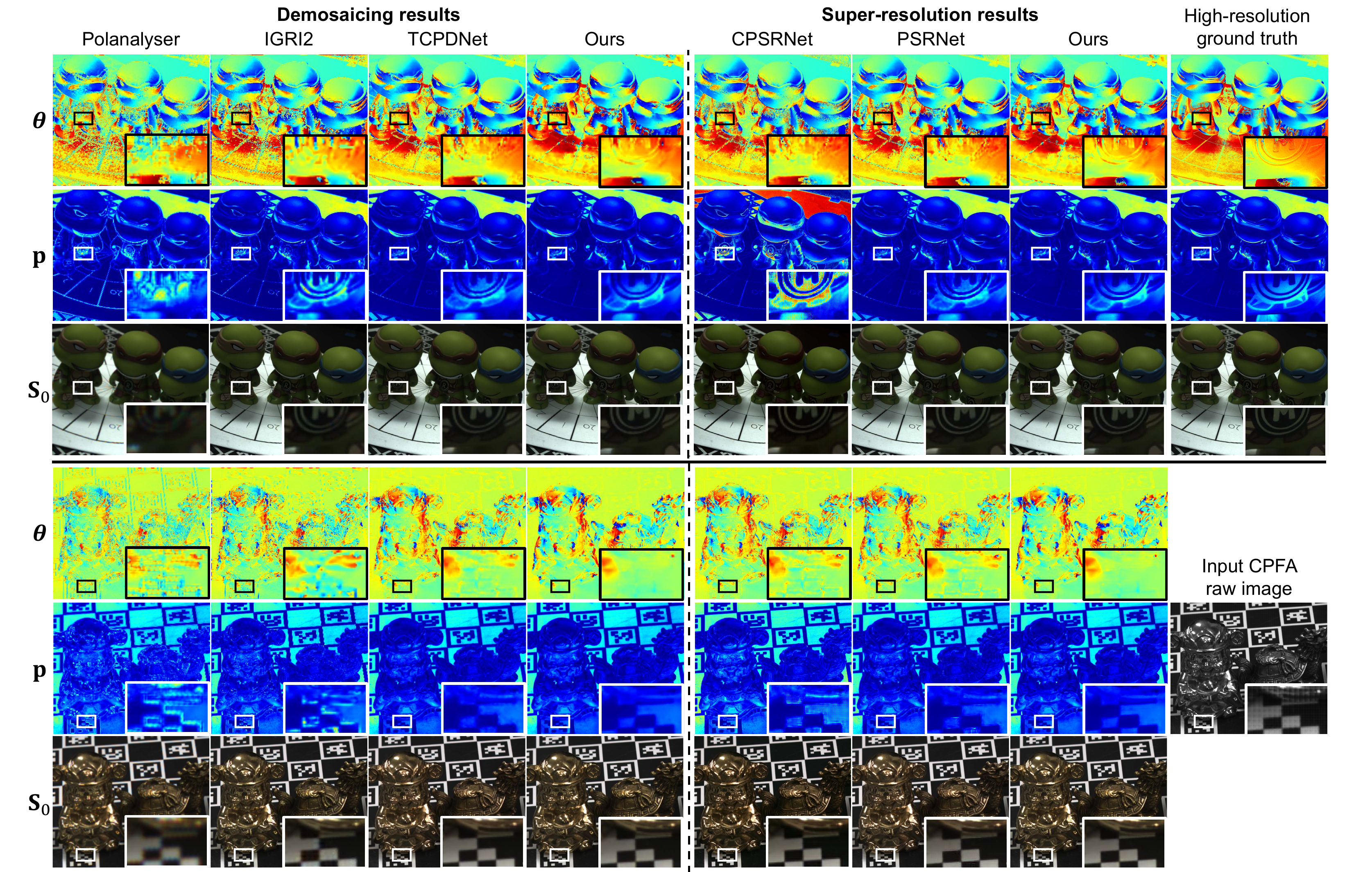}
    \caption{Qualitative comparisons on both synthetic (the top group) and real data (the bottom group) of both demosaicing and 4$\times$ SR tasks.} 
    \label{fig: SyntheticAndRealData}
\end{figure*}

\noindent\textbf{Loss function.} We design the loss function for both demosaicing and SR rounds as $L = \lambda_1 L_\text{img} + \lambda_2 L_\text{Stokes} + \lambda_3 L_\text{pol}$, where $L_\text{img}$ is the image loss aiming to ensure the pixel accuracy in the image domain, $L_\text{Stokes}$ is the Stokes loss aiming to preserve the continuity in the Stokes domain, and $L_\text{pol}$ is the polarization loss aiming to enforce the physical correctness of the DoP and AoP, $\lambda_{1,2,3}$ are set to be 1.0, 10.0, and 10.0 respectively. Here, we only detail each loss term in the demosaicing round, and the SR round could be similar (just replace the variables with the corresponding HR counterparts). For the second stage ($g$), $L_\text{img}$ can be written as $L_\text{img} = L_1(\mathbf{I}_{\alpha_1}+\mathbf{I}_{\alpha_3}, \mathbf{I}_{\alpha_2}+\mathbf{I}_{\alpha_4}) + L_\text{grad}(\mathbf{I}_{\alpha_{1,2,3,4}}, \mathbf{I}_{\alpha_{1,2,3,4}}^{\text{gt}})$, where $L_1$ and $L_\text{grad}$ denote the $\ell_1$ loss and gradient loss respectively, the superscript gt labels the ground truth throughout this paper. Here, $L_1(\mathbf{I}_{\alpha_1}+\mathbf{I}_{\alpha_3}, \mathbf{I}_{\alpha_2}+\mathbf{I}_{\alpha_4})$ aims to adjust the numerical relationship among $\mathbf{I}_{\alpha_{1,2,3,4}}$ since $\mathbf{I}_{\alpha_1}+\mathbf{I}_{\alpha_3} = \mathbf{I}_{\alpha_2}+\mathbf{I}_{\alpha_4}$ always holds for polarized images. $L_\text{Stokes}$ can be written as $L_\text{Stokes} = L_\text{grad}(\mathbf{S}_0, \mathbf{S}_0^{\text{gt}}) + L_1(\mathbf{S}_{1,2}, \mathbf{S}_{1,2}^{\text{gt}})$. $L_\text{pol}$ can be written as $L_\text{pol} = L_1(\mathbf{p}, \mathbf{p}^{\text{gt}}) + L_1(\bm{\theta}, \bm{\theta}^{\text{gt}})$. For the first stage ($f$), we use $\mathbf{T}_{\alpha_{1,2,3,4}}^{\text{gt}}$ (the half-resolution version of $\mathbf{I}_{\alpha_{1,2,3,4}}^{\text{gt}}$) along with the corresponding Stokes parameters, DoP, and AoP for supervision.

\noindent\textbf{Training strategy.} Our PIDSR is implemented using PyTorch and trained on an NVIDIA A800 GPU. For both demosaicing and SR, we train the two stages $f$ and $g$ for 100 epochs simultaneously in total, with a learning rate of 0.005. We use Adam optimizer \cite{kingma2014adam} for optimization.

\section{Experiment}
\subsection{Evaluation}
Since existing public datasets are insufficient for the setting of our PIDSR, we generate a synthetic dataset\footnote{Details about our dataset can be found in the supplementary material.} for evaluation. As for the demosaicing performance evaluation, we compare our PIDSR with three state-of-the-art PID methods Polanalyser \cite{maeda2019polanalyser}, IGRI2 \cite{morimatsu2021monochrome}, and TCPDNet \cite{nguyen2022two}; as for the SR performance evaluation, we compare our PIDSR with the only existing two PISR methods PSRNet \cite{hu2023polarized} and CPSRNet \cite{yu2023color}. Here, since PSRNet \cite{hu2023polarized} is initially designed for grayscale polarized images, we make slight modifications on it to allow it to accept the color polarized images. Besides, since the compared PISR methods \cite{hu2023polarized, yu2023color} can only take the polarized images (instead of the CPFA raw image) as input, we provide them the demosaicing results from TCPDNet \cite{nguyen2022two} (which achieves the best performance among the compared PID methods \cite{maeda2019polanalyser, morimatsu2021monochrome, nguyen2022two}). Note that all compared methods based on deep-learning \cite{nguyen2022two, hu2023polarized, yu2023color} are retrained on our dataset for a fair comparison. As the compared methods do, we not only evaluate the quality of $\mathbf{I}_{\alpha_{1,2,3,4}}$ (for demosaicing task) and $\mathbf{I}^\text{HR}_{\alpha_{1,2,3,4}}$ (for SR task), but also $\mathbf{p}$, $\bm{\theta}$, $\mathbf{S}_0$ (for demosaicing task) and $\mathbf{p}^\text{HR}$, $\bm{\theta}^\text{HR}$, $\mathbf{S}_0^\text{HR}$ (for SR task).

We evaluate the results quantitatively on synthetic data using: Mean Angular Error (MAE), Peak Signal-to-Noise Ratio (PSNR), and Structural Similarity Index Measure (SSIM). Here, MAE (lower values indicating better performance) is exclusively used to evaluate angular variables ($\bm{\theta}$ and $\bm{\theta}^\text{HR}$), while PSNR and SSIM are applied to the remaining variables. Results are shown in \tref{tab: SyntheticData}, where our framework consistently outperforms the compared methods on all metrics in both demosaicing and SR tasks. Visual quality comparisons on both synthetic and real data are shown in \fref{fig: SyntheticAndRealData}\footnote{Additional results can be found in the supplementary material.}. From the results, we can see that our PIDSR can produce more accurate DoP and AoP, while the compared methods suffer from severe artifacts (\eg, broken edges and discontinuity).

\begin{table}[t]
    \centering
    \caption{Quantitative evaluation results of ablation study.}
    \label{tab: AblationStudy}
    \resizebox{0.48\textwidth}{!}{
        \large
        \begin{tabular}{lcccccccc}
            \toprule
            {Metric} & \multicolumn{2}{c}{PSNR$\uparrow$/SSIM$\uparrow$} & {MAE$\downarrow$} \\
            \cmidrule[0.5pt](rl){1-1}
            \cmidrule[0.5pt](rl){2-3}
            \cmidrule[0.5pt](rl){4-4}
            {\large Demosaicing} & $\mathbf{S}_0$ & $\mathbf{p}$ & $\bm{\theta}$ \\
            \midrule
            Sequential $\mathcal{D}$ and $\uparrow$ & 32.32/0.9134 & 23.78/0.6661 & 19.2907 \\
            Single-stage pipeline & 34.61/0.9426 & 27.95/0.7406 & 38.2174 \\
            Without SFI blocks & 37.18/0.9612 & 32.73/0.8392 & 13.1242 \\
            Ours (demosaicing only)$\rightarrow$PSRNet \cite{hu2023polarized} & \textbf{40.24/0.9780} & \textbf{33.33/0.8447} & \textbf{12.2383} \\
            TCPDNet \cite{nguyen2022two}$\rightarrow$ ours (SR only) & 38.65/0.9669 & 32.26/0.8262 & 13.1881 \\
            \textbf{Our complete PIDSR} & \textbf{40.24/0.9780} & \textbf{33.33/0.8447} & \textbf{12.2383} \\
            \bottomrule
            \toprule
            {\large Super Resolution} & $\mathbf{S}_0^\text{HR}$ (2$\times$) & $\mathbf{p}^\text{HR}$ (2$\times$) & $\bm{\theta}^\text{HR}$ (2$\times$)\\
            \midrule
            Sequential $\mathcal{D}$ and $\uparrow$ & 32.16/0.8965 & 20.07/0.6225 & 21.8965 \\
            Single-stage pipeline & 34.35/0.9278 & 28.10/0.7584 & 38.2203 \\
            Without SFI blocks & 36.35/0.9458 & 31.65/0.8322 & 14.1309 \\
            Ours (demosaicing only)$\rightarrow$PSRNet \cite{hu2023polarized} & 36.81/0.9513 & 32.68/0.8412 & 12.5958 \\
            TCPDNet \cite{nguyen2022two}$\rightarrow$ ours (SR only) & 36.83/0.9457 & 32.19/0.8308 & 13.1530 \\
            \textbf{Our complete PIDSR} & \textbf{37.44/0.9553} & \textbf{32.97/0.8438} & \textbf{12.3520} \\
            \bottomrule
        \end{tabular}
    }
    \vspace{-2mm}
\end{table}

\subsection{Ablation study}
We conduct several ablation studies in \tref{tab: AblationStudy} to verify the validity of each design choice. First, we show the significance of our PIDSR framework design that formulates $\mathcal{D}^\uparrow$ as complementary demosaicing and SR, by comparing with an alternative design that naively formulates $\mathcal{D}^\uparrow$ as a combination of $\mathcal{D}$ and $\uparrow$ using the same network architecture (Sequential $\mathcal{D}$ and $\uparrow$). The performance degenerates, since such a naive pipeline would result in error accumulation. Next, we verify the necessity of our two-stage pipeline, by comparing to a single-stage pipeline that does not explicitly reconstruct spatial-physical coherence under the same PIDSR framework (Single-stage pipeline). The results are not that good since the still remaining spatial discontinuity and disrupted physical correlation would have negative impact on resolution enhancement. Then, we validate the effectiveness of our Stokes-aided neural network, by substituting the SFI blocks with original convolution blocks (Without SFI blocks). We find that it does not perform well since it cannot make full use of the Stokes-domain information to preserve the polarization properties. Finally, we also compare with two different hybrid baselines that feed our demosaicing results into PSRNet \cite{hu2023polarized} for SR (Ours (demosaicing only)$\rightarrow$PSRNet \cite{hu2023polarized}), and feed the demosaicing results of TCPDNet \cite{nguyen2022two} into our PIDSR for SR (TCPDNet \cite{nguyen2022two}$\rightarrow$ ours (SR only)), respectively. We can see that our complete PIDSR achieves the first performance.

\begin{figure}[t]
    \centering
    \includegraphics[width=1.0\linewidth]{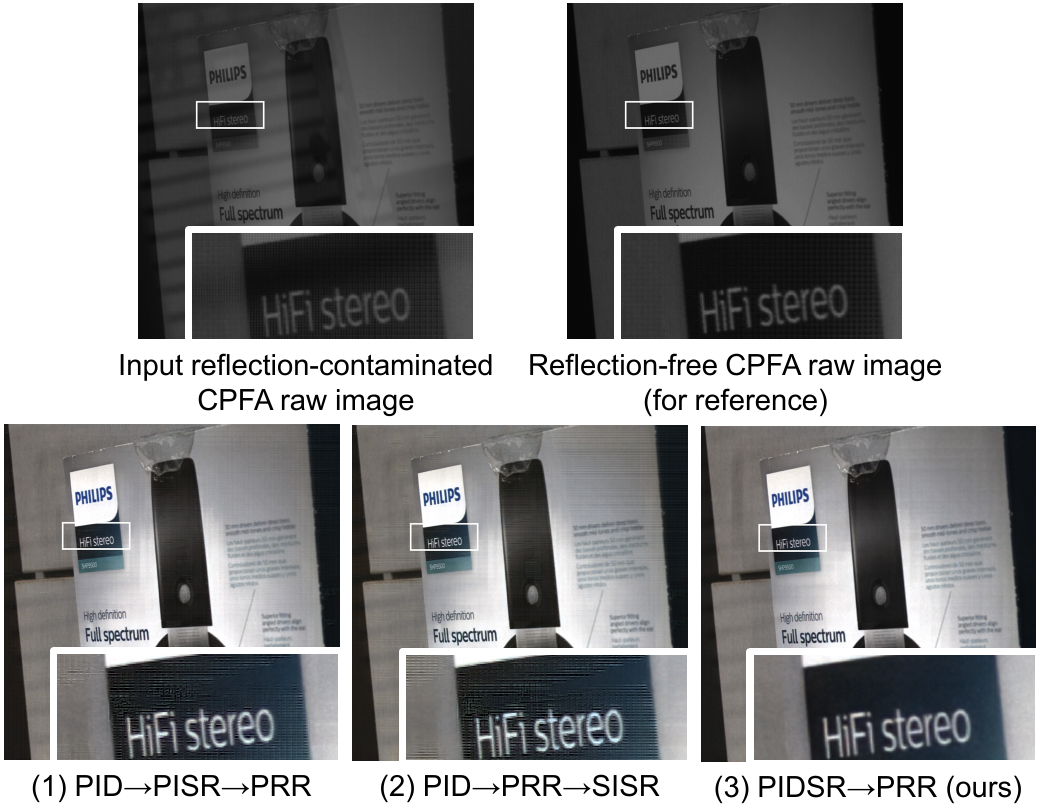}
    \caption{Results of polarization-based reflection removal. Our method is not influenced by the zigzag artifacts. Please zoom-in for better details.}
    \label{fig: Application}
\end{figure}

\subsection{Application}
To show that our PIDSR can be beneficial to downstream polarization-based vision applications, we take polarization-based reflection removal (PRR, which takes reflection-contaminated polarized images as input and outputs reflection-removed unpolarized images) as an example, and try to obtain a reflection-removed unpolarized image from a reflection-contaminated CPFA raw image captured by a polarization camera. To achieve it, the following approaches could be used: (1) ``PID$\rightarrow$PISR$\rightarrow$PRR'': performing PID and PISR sequentially on the CPFA raw image, then performing reflection removal; (2) ``PID$\rightarrow$PRR$\rightarrow$SISR'': performing PID on the CPFA raw image first, then performing reflection removal, and performing single image super-resolution (SISR) in the end; (3) ``PIDSR$\rightarrow$PRR'': performing our PIDSR on the CPFA raw image first, then performing reflection removal. Here, the SR scale is 2, and the involved PRR, PID, PISR, and SISR methods are selected to be RSP \cite{lyu2019reflection}, TCPDNet \cite{nguyen2022two}, PSRNet \cite{hu2023polarized}, and OmniSR \cite{wang2023omni} respectively. Visual comparisons are shown in \fref{fig: Application}, where we can see the result from PIDSR$\rightarrow$PRR (ours) contains more detailed textures and less reflection contamination.

\section{Conclusion}
We propose PIDSR, a joint framework that performs complementary polarized image demosaicing and super-resolution. By carefully designing a two-stage recurrent pipeline to fundamentally reduce the level of error and a Stokes-aided neural network to preserve the polarization properties, our PIDSR can robustly obtain HR polarized images with more accurate polarization-related parameters such as the DoP and AoP from a CPFA raw image in a direct manner.

\noindent\textbf{Limitations.} Since our PIDSR is specifically designed to process a single CPFA raw image, it is unsuitable for reconstructing a polarized video. Additionally, it cannot handle CFA raw images, as it requires the Stokes parameters as input, which are unavailable in this setting.


\section*{Acknowledgment} This work was supported by Hebei Natural Science Foundation Project No. F2024502017, Beijing-Tianjin-Hebei Basic Research Funding Program No. 242Q0101Z, National Natural Science Foundation of China (Grant No. 62472044, U24B20155, 62136001, 62088102, 62225601, U23B2052), Beijing Municipal Science \& Technology Commission, Administrative Commission of Zhongguancun Science Park (Grant No. Z241100003524012), the JST-Mirai Program Grant Number JPMJM123G1. BUPT and PKU affiliated authors thank openbayes.com for providing computing resource.

\clearpage
{
    \small
    \bibliographystyle{ieeenat_fullname}
    \bibliography{main}
}

\end{document}